\documentclass[sigconf]{aamas}  
\usepackage{booktabs}
\usepackage{lipsum}
\usepackage{graphicx}
\usepackage{natbib}
\usepackage{color}
\usepackage{algorithm}
\usepackage{algorithmic}
\usepackage[scaled=.8]{beramono} 
\usepackage{amsmath}
\usepackage{amssymb}
\newcommand{\xmark}{\ding{55}}%
\usepackage{subcaption}
\usepackage{pifont}
\usepackage[utf8]{inputenc} 
\usepackage[T1]{fontenc}    
\usepackage{url}            
\usepackage{booktabs}       
\usepackage{amsfonts}       
\usepackage{nicefrac}       
\usepackage{microtype}      
\usepackage{multicol}
\graphicspath{{./aamas_2018/figs/}}

\DeclareMathOperator*{\softmax}{softmax}
\newcommand{\xatt}{x_{\oplus}}
 
\newcommand{\bi}{\begin{itemize}}
\newcommand{\ei}{\end{itemize}}
\newcommand{\be}{\begin{enumerate}}
\newcommand{\ee}{\end{enumerate}}

\setcopyright{ifaamas}  
\acmDOI{}  
\acmISBN{}  
\acmConference[AAMAS'18]{Proc.\@ of the 17th International Conference on Autonomous Agents and Multiagent Systems (AAMAS 2018)}{July 10--15, 2018}{Stockholm, Sweden}{M.~Dastani, G.~Sukthankar, E.~Andr\'{e}, S.~Koenig (eds.)}  
\acmYear{2018}  
\copyrightyear{2018}  
\acmPrice{}  

\begin{document}

\title{Crossmodal Attentive Skill Learner}  

\author{Shayegan Omidshafiei}
\affiliation{%
  \institution{Massachusetts Institute of Technology}
  \city{Cambridge} 
  \state{Massachusetts} 
}
\email{shayegan@mit.edu}

\author{Dong-Ki Kim}
\affiliation{%
	\institution{Massachusetts Institute of Technology}
	\city{Cambridge} 
	\state{Massachusetts} 
}
\email{dkkim93@mit.edu}

\author{Jason Pazis}
\authornote{Work done prior to Amazon involvement of the author, and does not reflect views of the Amazon company.}
\affiliation{%
	\institution{Amazon Alexa}
	\city{Cambridge} 
	\state{Massachusetts} 
}
\email{pazisj@amazon.com}

\author{Jonathan P.~How}
\affiliation{%
	\institution{Massachusetts Institute of Technology}
	\city{Cambridge} 
	\state{Massachusetts} 
}
\email{jhow@mit.edu}

\renewcommand{\shortauthors}{S. Omidshafiei et al.}

\begin{abstract}
    	This paper introduces the Crossmodal Attentive Skill Learner (CASL), integrated with the recently-introduced Asynchronous Advantage Option-Critic (A2OC) architecture \citep{harb2017waiting} to enable hierarchical reinforcement learning across multiple sensory inputs. 
    	We provide concrete examples where the approach not only improves performance in a single task, but accelerates transfer to new tasks. 
    	We demonstrate the attention mechanism anticipates and identifies useful latent features, while filtering irrelevant sensor modalities during execution. 
		We modify the Arcade Learning Environment \citep{bellemare13arcade} to support audio queries, and conduct evaluations of crossmodal learning in the Atari 2600 games H.E.R.O. and Amidar.
		Finally, building on the recent work of \citet{babaeizadeh2017ga3c}, we open-source a fast hybrid CPU-GPU implementation of CASL.\footnote{Code available at \url{https://github.com/shayegano/CASL}} 
\end{abstract}


\maketitle


\section{Introduction}
Intelligent agents should be capable of disambiguating local sensory streams to realize long-term goals. 
In recent years, the combined progress of computational capabilities and algorithmic innovations has afforded reinforcement learning (RL) \citep{sutton1998reinforcement} approaches the ability to achieve this desiderata in impressive domains, exceeding expert-level human performance in tasks such as Atari and Go \citep{mnih2015human,silver2017mastering}.
Nonetheless, many of these algorithms thrive primarily in well-defined mission scenarios learned in isolation from one another; such monolithic approaches are not sufficiently scalable missions where goals may be less clearly defined, and sensory inputs found salient in one domain may be less relevant in another. 

How should agents learn effectively in domains of high dimensionality, where tasks are durative, agents receive sparse feedback, and sensors compete for limited computational resources? One promising avenue is hierarchical reinforcement learning (HRL), focusing on problem decomposition for learning transferable skills. Temporal abstraction enables exploitation of domain regularities to provide the agent hierarchical guidance in the form of options or sub-goals \citep{sutton1999between,kulkarni2016hierarchical}.
Options help agents improve learning by mitigating scalability issues in long-duration missions, by reducing the effective number of decision epochs.
In the parallel field of supervised learning, temporal dependencies have been captured proficiently using attention mechanisms applied to encoder-decoder based sequence-to-sequence models \mbox{\citep{bahdanau2014neural,luong2015effective}}.
\emph{Attention} empowers the learner to focus on the most pertinent stimuli and capture longer-term correlations in its encoded state, for instance to conduct neural machine translation or video captioning \citep{yeung2015every,yang2016hierarchical}.
Recent works also show benefits of spatio-temporal attention in RL \citep{mnih2014recurrent,sorokin2015deep}.

One can interpret the above approaches as conducting dimensionality reduction, where the target dimension is \emph{time}. 
In view of this insight, this paper proposes an RL paradigm exploiting hierarchies in the dimensions of \emph{time} and \emph{sensor modalities}. Our aim is to learn rich skills that attend to and exploit pertinent crossmodal (multi-sensor) signals at the appropriate moments. The introduced crossmodal skill learning approach largely benefits an agent learning in a high-dimensional domain (e.g., a robot equipped with many sensors). Instead of the expensive operation of processing and/or storing data from all sensors, we demonstrate that our approach enables such an agent to focus on important sensors; this, in turn, leads to more efficient use of the agent's limited computational and storage resources (e.g., its finite-sized memory). 

In this paper, we focus on combining two sensor modalities: audio and video. While these modalities have been previously used for supervised learning \citep{ngiam2011multimodal}, to our knowledge they have yet to be exploited for crossmodal skill learning. We provide concrete examples where the proposed HRL approach not only improves performance in a single task, but accelerates transfer to new tasks. 
We demonstrate the attention mechanism anticipates and identifies useful latent features, while filtering irrelevant sensor modalities during execution. 
We also show preliminary results in the Arcade Learning Environment \citep{bellemare13arcade}, which we modified to support audio queries. 
In addition, we provide insight into how our model functions internally by analyzing the interactions of attention and memory.
Building on the recent work of \citet{babaeizadeh2017ga3c}, we open-source a fast hybrid CPU-GPU implementation of our framework. Finally, note that despite this paper's focus on audio-video sensors, the framework presented is general and readily applicable to additional sensory inputs. 
\section{Related Work}
Our work is most related to crossmodal learning approaches that take advantage of multiple input sensor modalities.
Fusion of multiple modalities or sources of information is an active area of research. 
Works in the diverse domains of sensor fusion in robotics \cite{lynen13fusion, chambers14fusion, nobili17fusion}, audio-visual fusion \cite{ngiam2011multimodal, srivastava14fusion, beal02fusion, bengio02fusion}, and image-point cloud fusion \cite{cadena14fusion, alvis17fusion} have shown that models utilizing multiple modalities tend to outperform those learned from unimodal inputs.
In general, approaches for multimodal fusion can be broadly classified depending on the means of integration of the various information sources.
Filtering-based frameworks (e.g., the Extended Kalman Filter) are widely used to combine multi-sensor readings in the robotics community \cite{lynen13fusion, chambers14fusion, nobili17fusion}.
In machine learning, approaches based on graphical models \cite{beal02fusion, bengio02fusion} and conditional random fields \cite{lafferty01crf} have been used to integrate multimodal features \cite{cadena14fusion, alvis17fusion}.
More recently, deep learning-based approaches that learn a representations of features across multiple modalities have been introduced \cite{ngiam2011multimodal, srivastava14fusion, kiros14fusion, eitel15iros}. 

A variety of attentive mechanisms have been considered in recent works, primarily in application to supervised learning. Temporal attention mechanisms have been successfully applied to the field of neural machine translation, for instance in \cite{bahdanau2014neural,luong2015effective}, where encoder-decoder networks work together to transform one sequence to another. Works also exist in multimodal attention for machine translation, using video-text inputs \cite{caglayan2016multimodal}. Spatial attention models over image inputs have also been combined with Deep Recurrent Q-Networks \cite{hausknecht2015deep} for RL \cite{sorokin2015deep}. Works have also investigated spatially-attentive agents trained via RL to conduct image classification \cite{mnih2014recurrent,ba2014multiple}. 
As we later demonstrate, the crossmodal attention-based approach used in this paper enables filtering of irrelevant sensor modalities, leading to improved learning and more effective use of the agent's memory.

There exists a large body of HRL literature, targeting both fully and partially-observable domains. Our work leverages the options framework \cite{sutton1999between}, specifically the recent Asynchronous Advantage Option-Critic (A2OC) \cite{harb2017waiting} algorithm, to learn durative skills. HRL is an increasingly-active field, with a number of recent works focusing on learning human-understandable and/or intuitive skills. In \cite{andreas2016modular}, annotated descriptors of policies are used to enable multitask RL using options. Multitask learning via parameterized skills is also considered in \cite{da2012learning}, where a classifier and regressor are used to, respectively, identify the appropriate policy to execute, then map to the appropriate parameters of said policy. Construction of \emph{skill chains} is introduced in \cite{konidaris2009skill} for learning in continuous domains. In \cite{machado2017laplacian}, option discovery is conducted through eigendecomposition of MDP transition matrices, leading to transferable options that are agnostic of the task reward function. FeUdal Networks \cite{vezhnevets2017feudal} introduce a two-level hierarchy, where a manager defines sub-goals, and a worker executes primitive actions to achieve them. A related HRL approach is also introduced in \cite{kulkarni2016hierarchical}, where sub-goals are hand-crafted by a domain expert. Overall, the track record of hierarchical approaches for multitask and transfer learning leads us to use them as a basis for the proposed framework, as our goal is to learn scalable policies over high-volume input streams.

While the majority of works utilizing multi-sensory and attentive mechanisms focus on supervised learning, our approach targets RL. Specifically, we introduce an HRL framework that combines crossmodal learning, attentive mechanisms, and temporal abstraction to learn durative skills. 
\section{Background}
This section summarizes Partially Observable Markov Decision Processes (POMDPs) and options, which serve as foundational frameworks for our approach.

\subsection{POMDPs}\label{sec:POMDP}
This work considers an agent operating in a partially-observable stochastic environment, modeled as a POMDP $\langle \mathbb{S}, \mathbb{A}, \mathbb{O}, \mathcal{T}, \mathcal{O}, \mathcal{R}, \gamma \rangle$ \citep{kaelbling1998planning}. $\mathbb{S}$, $\mathbb{A}$, and $\mathbb{O}$ are, respectively, the state, action, and observation spaces. At timestep $t$, the agent executes action $a \in \mathbb{A}$ in state $s \in \mathbb{S}$, transitions to state $s' \sim \mathcal{T}(s, a, s')$, receives observation $o \sim \mathcal{O}(o,s',a)$, and reward $r_{t} = \mathcal{R}(s,a) \in \mathbb{R}$. The value of state $s$ under policy $\pi:Dist(\mathbb{S})\to\mathbb{A}$ is the expected return $V_{\pi}(s)=\mathbb{E}[\sum_{k=0}^{H}\gamma^{k}r_{t+k+1} | s_t = s ]$, given timestep $t$, horizon $H$, and discount factor $\gamma \in [0,1)$. The objective is to learn an optimal policy ${\pi}^*$, which maximizes the value. 

As POMDP agents only receive noisy observations of the latent state, policy $\pi$ typically maps from the agent's belief (distribution over states) to the next action. Recent work has introduced Deep Recurrent Q-Networks (DRQNs) \cite{hausknecht2015deep} for RL in POMDPs, leveraging recurrent Neural Networks (RNNs) that inherently maintain an internal state $h^{t}$ to compress input history until timestep $t$.

\subsection{Options}\label{sec:options}
The framework of \emph{options} provides an RL agent the ability to plan using temporally-extended actions \cite{sutton1999between}. Option $\mathcal{\omega}\in \Omega$ is defined by initiation set $\mathbb{I}\subseteq\mathbb{S}$, intra-option policy $\pi_{\omega}: \mathbb{S}\to \text{Dist}(\mathbb{A})$, and termination condition $\beta_{\omega}: \mathbb{S}\to[0,1]$. A policy over options $\pi_{\Omega}$ chooses an option among those that satisfy the initiation set. The selected option executes its intra-option policy until termination, upon which a new option is chosen. This process iterates until the goal state is reached. Recently, the Asynchronous Advantage Actor-Critic framework (A3C) \citep{mnih2016asynchronous} has been applied to POMDP learning in a computationally-efficient manner by combining parallel actor-learners and Long Short-Term Memory (LSTM) cells \citep{hochreiter1997long}. Asynchronous Advantage Option-Critic (A2OC) extends A3C and enables learning option-value functions, intra-option policies, and termination conditions in an end-to-end fashion \citep{harb2017waiting}. 
The option-value function models the value of state $s \in \mathbb{S}$ in option $\omega \in \Omega$,
\begin{equation}
Q_{\Omega}(s, \omega)=\sum\limits_{a} \pi_{\omega}(a|s) \Big(r(s,a) + \gamma\sum\limits_{s'}\mathcal{T}(s'|s,a)U(s', \omega)\Big),
\end{equation} 
where $a \in \mathbb{A}$ is a primitive action and $U(s', \omega)$ represents the option utility function,
\begin{equation}
U(s', \omega)=(1-\beta_{\omega}(s'))Q_{\Omega}(\omega,s')+\beta_{\omega}(s')(V_{\Omega}(s')-c).
\end{equation} 
A2OC introduces deliberation cost, $c$, in the utility function to address the issue of options terminating too frequently. Intuitively, the role of $c$ is to impose an added penalty when options terminate, enabling regularization of termination frequency. 
The value function over options, $V_{\Omega}$, is defined,
\begin{equation}\label{}
V_{\Omega}(s')=\sum\limits_{\omega} \pi_{\Omega}(\omega|s') Q_{\Omega}(\omega,s'),
\end{equation} 
where $\pi_{\Omega}$ is the policy over options (e.g., an epsilon-greedy policy over $Q_{\Omega}$). 
Assuming use of a differentiable representation, option parameters can be learned using gradient descent.
Readers are referred to \cite{harb2017waiting} for more details.
\section{Approach}
\begin{figure*}[t]
	\centering
	\includegraphics[width=0.6\linewidth, trim={0 0.2cm 0 0.3cm},clip]{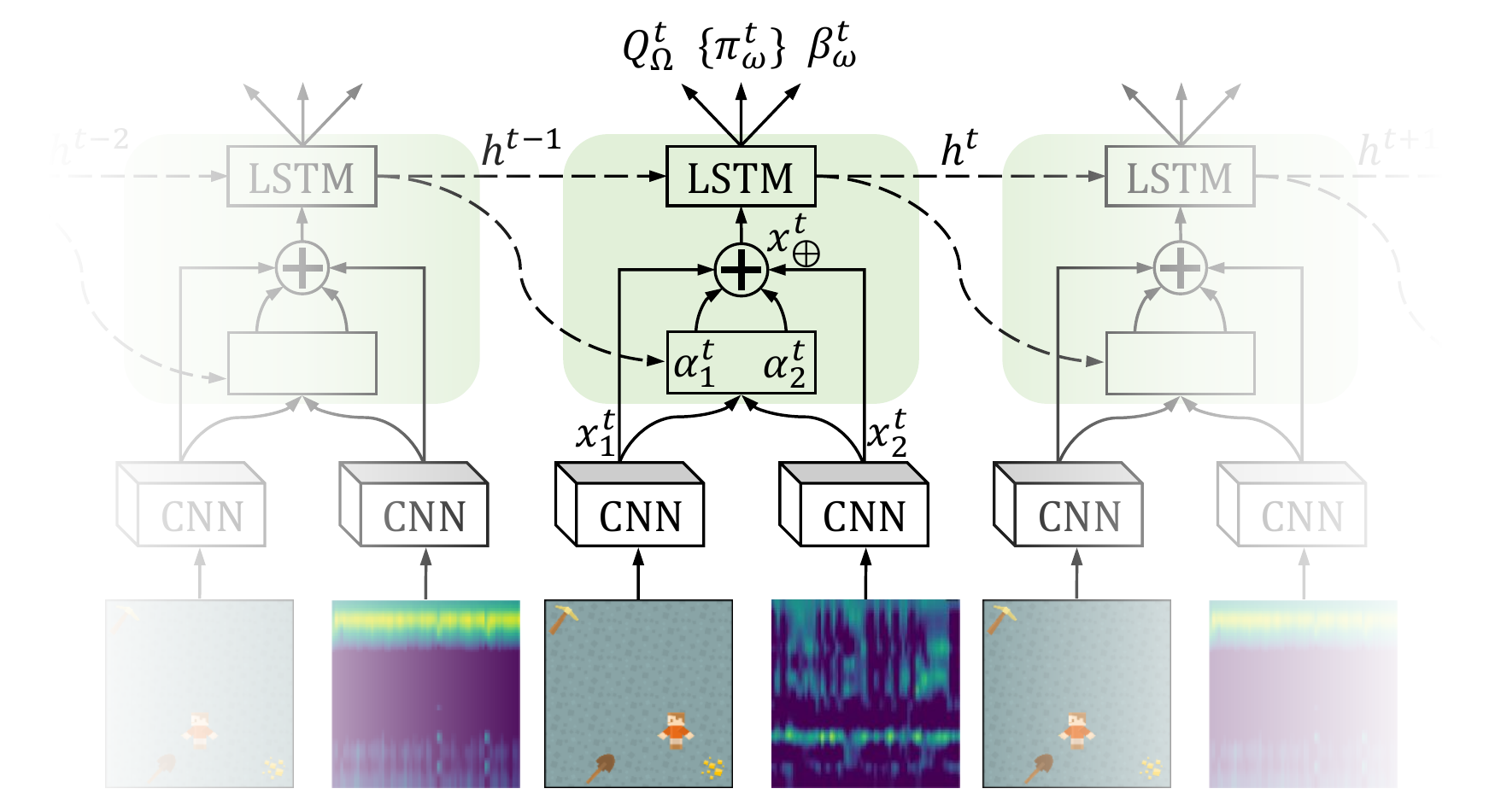}
	\caption{CASL network architecture enables attention-based learning over multi-sensory inputs. Green highlighted region indicates crossmodal attention LSTM cell, trained via backpropagation through time.}
	\label{fig:network_architecture}
\end{figure*}

Our goal is to design a mechanism that enables the learner to modulate high-dimensional sensory inputs, focusing on pertinent stimuli that may lead to more efficient skill learning. This section presents motivations behind attentive skill learning, then introduces the proposed framework.

\subsection{Attentive Mechanisms}\label{sec:attentive_mechanisms}
Before presenting the proposed architecture, let us first motivate our interests towards attentive skill learning. One might argue that the combination of deep learning and RL already affords agents the representation learning capabilities necessary for proficient decision-making in high-dimensional domains; i.e., why the need for crossmodal attention? 

Our ideas are motivated by the studies in behavioral neuroscience that suggest the interplay of attention and choice bias humans' value of information during learning, playing a key factor in solving tasks with high-dimensional information streams  \citep{leong2017dynamic}. Works studying learning in the brain also suggest a natural pairing of attention and hierarchical learning, where domain regularities are embedded as priors into skills and combined with attention to alleviate the curse of dimensionality \citep{niv2015reinforcement}. Works also suggest attention plays a role in the intrinsic curiosity of agents during learning, through direction of focus to regions predicted to have high reward \citep{mackintosh1975theory}, high uncertainty \citep{pearce1980model}, or both \citep{pearce2010two}. 

In view of these studies, we conjecture that crossmodal attention, in combination with HRL, improves representations of relevant environmental features that lead to superior learning and decision-making. Specifically, using crossmodal attention, agents combine internal beliefs with external stimuli to more effectively exploit multiple modes of input features for learning. As we later demonstrate, our approach captures temporal crossmodal dependencies, and enables faster and more proficient learning of skills in the domains examined.

\subsection{Crossmodal Attentive Skill Learner}\label{sec:CASL} 
We propose Crossmodal Attentive Skill Learner (CASL), a novel framework for HRL. One may consider many blueprints for integration of multi-sensory attention into the options framework. Our proposed architecture is primarily motivated by the literature that taxonomizes attention into two classes: \emph{exogeneous} and \emph{endogeneous}. The former is an involuntary mechanism triggered automatically by the inherent saliency of the sensory inputs, whereas the latter is driven by the intrinsic and possibly long-term goals, intents, and beliefs of the agent \citep{carrasco2011visual}. Previous attention-based neural architectures take advantage of both classes, for instance, to solve natural language processing problems \citep{vinyals2015grammar}; our approach follows a similar schema.

The CASL network architecture is visualized in \Cref{fig:network_architecture}. Let $M \in \mathbb{N}$ be the number of sensor modalities (e.g., vision, audio, etc.) and $x_{m}$ denote extracted features from the $m$-th sensor, where $m \in \{1,\ldots,M\}$. For instance, $x_{m}$ may correspond to feature outputs of a convolutional neural network given an image input. Given extracted features for all $M$ sensors at timestep $t$, as well as hidden state $h^{t-1}$, the proposed crossmodal attention layer learns the relative importance of each modality $\alpha^{t} \in \Delta^{M-1}$, where $\Delta^{M-1}$ is the $(M-1)$-simplex:
\begin{align}
	z^t &= \tanh\Big(\underbrace{\sum\limits_{m=1}^{M} (W_{m}^{T}x_{m}^{t} + b_{m})}_{\text{Exogeneous attention}} + \underbrace{\vphantom{\sum\limits_{m=1}^{M}} W_{h}^{T}h^{t-1} + b_{h}}_{\text{Endogeneous attention}}\Big)\label{eqn:attn_feats}\\
	\alpha^{t} &= \softmax\left(W_{z}^{T}z^{t}+b_{z}\right)\\
	\xatt &= \begin{cases}
	\sum_{m=1}^{M} \alpha_{m}^{t}x_{m}^{t}  &\text{(Summed attention)}\\[6pt]
	\left[(\alpha_{1}^{t}x_{1}^{t})^T, \ldots, (\alpha_{m}^{t}x_{m}^{t})^T\right]^T  &\text{(Concatenated attention)} \label{eqn:attn_attended_feat_concat}
	\end{cases}
\end{align}
Weight matrices $W_{m}$, $W_{h}$, $W_{z}$ and bias vectors $b_{m}$, $b_{h}$, $b_{z}$ are trainable parameters and nonlinearities are applied element-wise.  

Both exogeneous attention over sensory features $x_{m}^{t}$ and endogeneous attention over LSTM hidden state $h^{t-1}$ are captured in \Cref{eqn:attn_feats}.  The sensory feature extractor used in experiments consists of $3$ convolutional layers, each with $32$ filters of size $3\times3$, stride $2$, and ReLU activations. Attended features $\alpha_{m}^{t}x_{m}^{t}$ may be combined via summation or concatenation (\Cref{eqn:attn_attended_feat_concat}), then fed to an LSTM cell. The LSTM output captures temporal dependencies used to estimate option values, intra-option policies, and termination conditions ($Q_{\Omega}$, $\pi_{\omega}$, $\beta_{\omega}$ in \Cref{fig:network_architecture}, respectively),
\begin{align}
Q_{\Omega}(s,\omega)&=W_{Q,\omega}h^t+b_{Q,\omega},\\
\pi_{\omega}(a|s)&=\softmax(W_{\pi,\omega}h^t+b_{\pi,\omega}),\\
\beta_{\omega}(s)&=\sigma(W_{\beta,\omega}h^t+b_{\beta,\omega}),
\end{align}
where weight matrices $W_{\Omega,\omega}$, $W_{\pi,\omega}$, $W_{\beta,\omega}$ and bias vectors $b_{\Omega,\omega}$, $b_{\pi,\omega}$, $b_{\beta,\omega}$ are trainable parameters for the current option $\omega$, and $\sigma(\cdot)$ is the sigmoid function.
Network parameters are updated using gradient descent. Entropy regularization of attention outputs $\alpha^{t}$ was found to encourage exploration of crossmodal attention behaviors during training.
\section{Evaluation}

The proposed framework is evaluated on a variety of learning tasks with inherent reward sparsity and transition noise. We evaluate our approach in three domains: a door puzzle domain, a 2D-Minecraft like domain, and the Arcade Learning Environment \cite{bellemare13arcade}. These environments include challenging combinations of reward sparsity and/or complex audio-video sensory input modalities that may not always be useful to the agent. 
The first objective of our experiments is to analyze performance of CASL in terms of learning rate and transfer learning.
The second objective is to understand relationships between attention and memory mechanisms (as captured in the LSTM cell state). 
Finally, we modify the Arcade Learning Environment to support audio queries, and evaluate crossmodal learning in the Atari 2600 games H.E.R.O. and Amidar.

\subsection{Crossmodal Learning and Transfer}\label{sec:learning_and_transfer}
We first evaluate crossmodal attention in a sequential door puzzle game, where the agent spawns in a 2D world with two locked doors and a key at fixed positions. The key type is randomly generated, and its observable color indicates the associated door. The agent hears a fixed sound (but receives no reward) when adjacent to the key, and hears noise otherwise. The agent must find and pick up the key (which then disappears), then find and open the correct door to receive $+1$ reward ($\gamma = 0.99$). The game terminates upon opening of either door. The agent's sensory inputs $x_{m}^{t}$ are vision  and audio spectrogram. This task was designed in such a way that audio is not necessary to achieve the task -- the agent can certainly focus on learning a policy mapping from visual features to open the correct door. However, audio provides potentially useful signals that may accelerate learning, making this a domain of interest for analyzing the interplay of attention and sensor modalities.

\begin{figure}[t]
	\begin{subfigure}[t]{1\linewidth}
		\centering
		\includegraphics[width=1\linewidth]{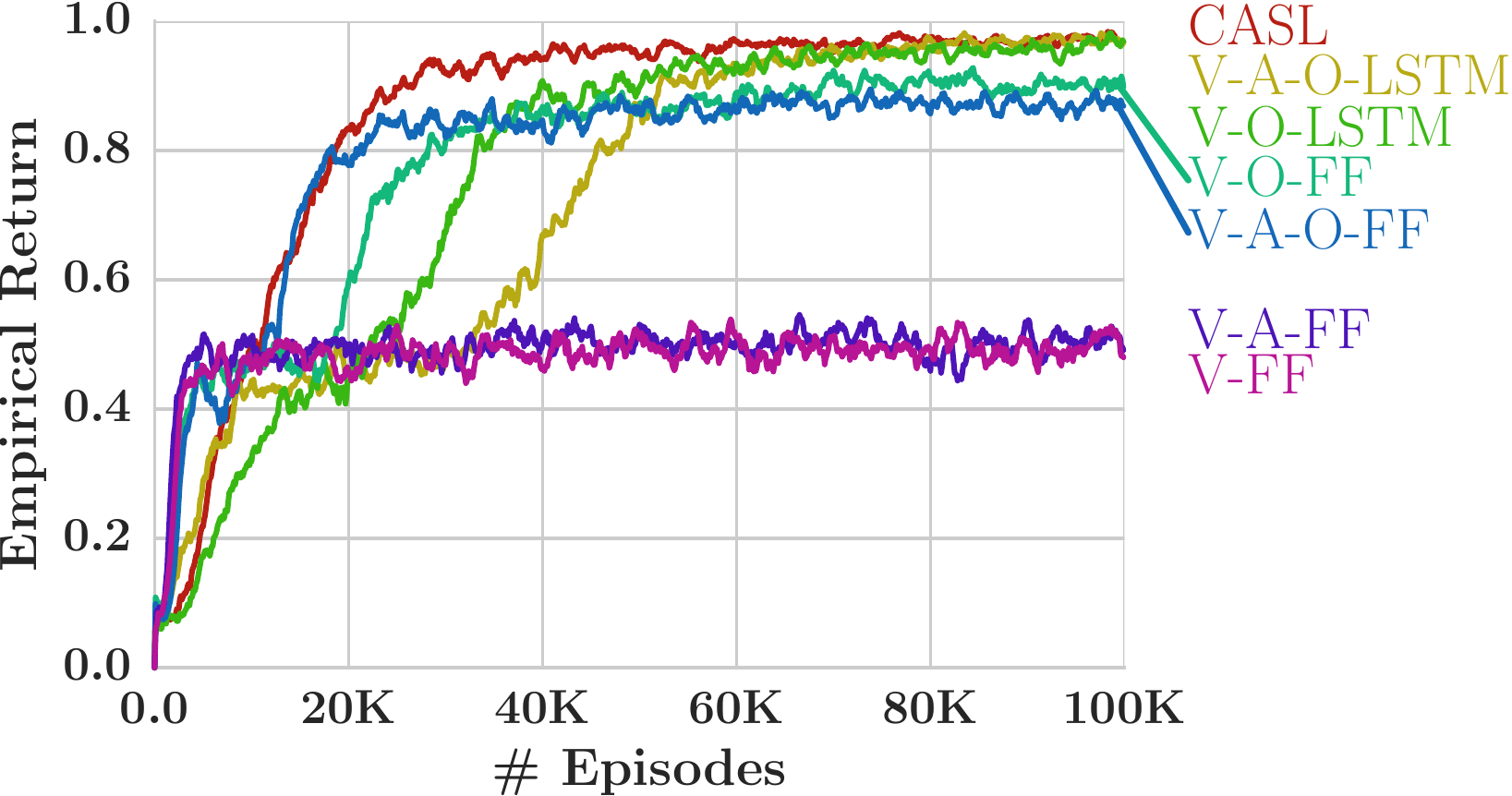}
		\caption{Door puzzle domain.}
		\label{fig:plot_mazeworld_comparisons}
	\end{subfigure}
	\\
	\begin{subfigure}[t]{1\linewidth}
		\centering
		\includegraphics[width=1\linewidth]{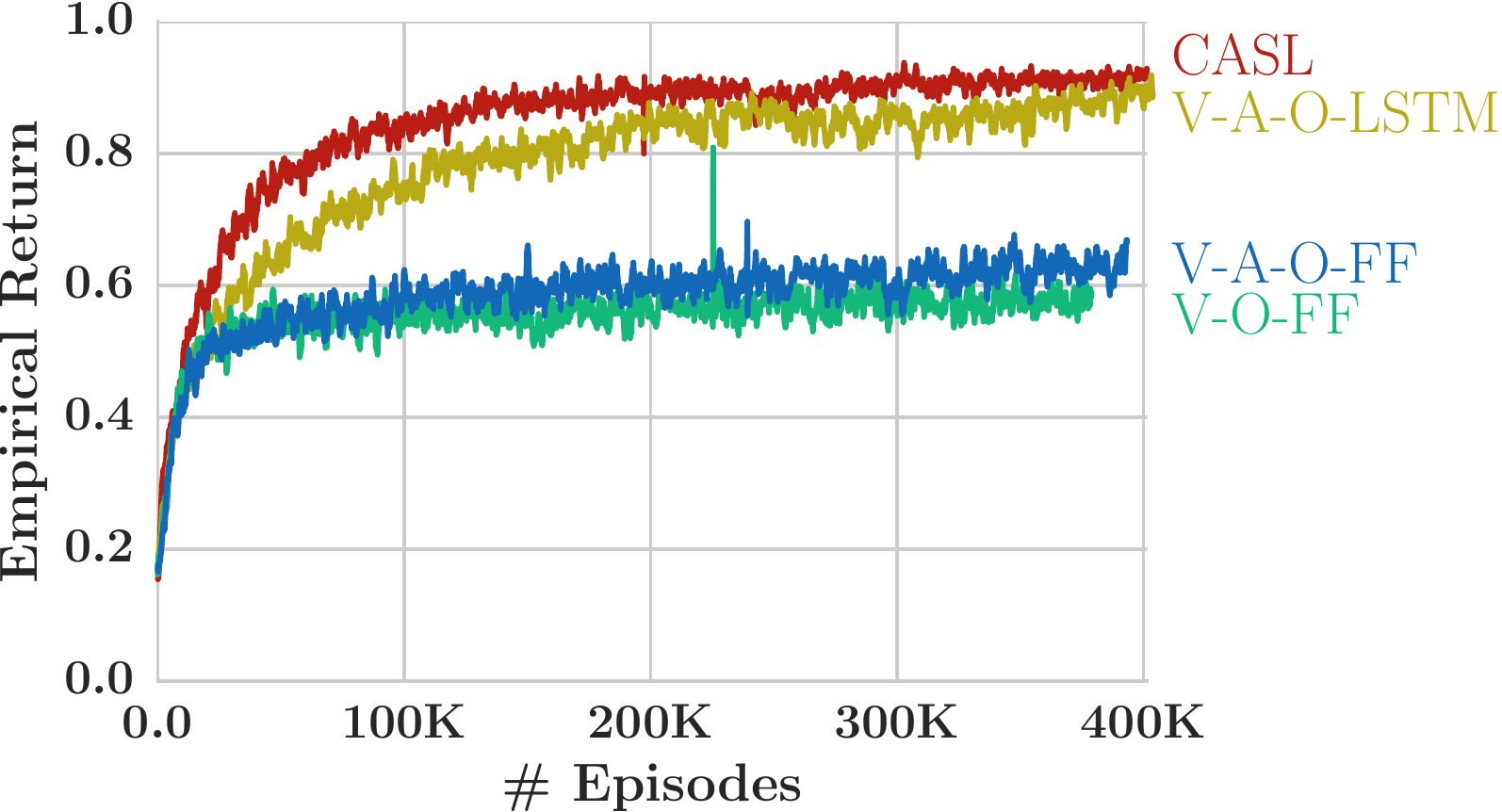}
		\caption{Door puzzle transfer learning.}
		\label{fig:transfer_data_final}
	\end{subfigure}
	\caption{CASL improves learning rate compared to other networks. Abbreviations: \textbf{V}: video, \textbf{A}: audio, \textbf{O}: options, \textbf{FF} feedforward net, \textbf{LSTM}: Long Short-Term Memory net.}
\end{figure}

\paragraph{Attention Improves Learning Rate} \Cref{fig:plot_mazeworld_comparisons} shows ablative training results for several network architectures. The three LSTM-based skill learners (including CASL) converge to the optimal value. Interestingly, the network that ignores audio inputs (V-O-LSTM) converges faster than its audio-enabled counterpart (V-A-O-LSTM), indicating the latter is overwhelmed by the extra sensory modality. Introduction of crossmodal attention enables CASL to converge faster than all other networks, using roughly half the training data of the others. The feedforward networks all fail to attain optimal value, with the non-option cases (V-A-FF and V-FF) repeatedly opening one door due to lack of memory of key color. Notably, the option-based feedforward nets exploit the option index to implicitly remember the key color, leading to higher value. Interplay between explicit memory mechanisms and use of options as pseudo-memory may be an interesting line of future work. 

\paragraph{Attention Accelerates Transfer} We also evaluate crossmodal attention for transfer learning (\Cref{fig:transfer_data_final}), using the more promising option-based networks. The door puzzle domain is modified to randomize the key position, with pre-trained options from the fixed-position variant used for initialization. All networks benefit from an  empirical return jumpstart of nearly 0.2 at the beginning of training, due to skill transfer. Once again, CASL converges fastest, indicating more effective use of the available audio-video data. While the asymptotic performance of CASL is only slightly higher than the V-A-O-LSTM network, the reduction in number of samples needed to achieve a high score (e.g., after 100K episodes) makes it advantageous for domains with high sampling cost.

\begin{figure}[t]
	\begin{subfigure}[t]{0.3\linewidth}
		\centering
		\includegraphics[width=0.95\linewidth,trim={0 0 0 0},clip]{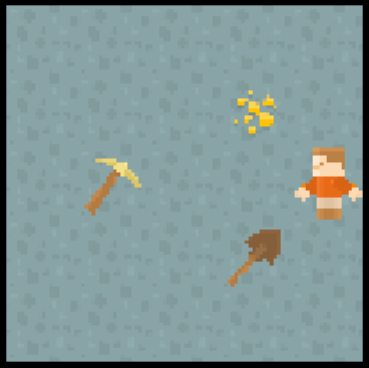}
		\caption{Gold ore should be mined with the pickaxe.}
		\label{fig:sdworld_hard_gold}
	\end{subfigure}
	\hfill	
	\begin{subfigure}[t]{0.3\linewidth}
		\centering
		\includegraphics[width=0.95\linewidth,trim={0 0 0 0},clip]{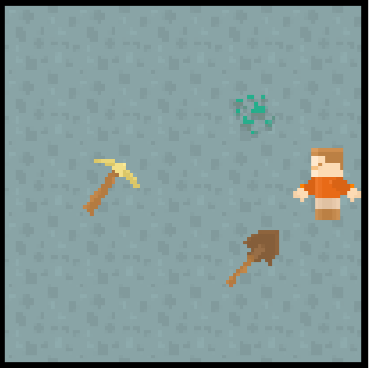}
		\caption{Iron ore should be mined with the shovel.}
		\label{fig:sdworld_hard_iron}
	\end{subfigure}
	\hfill
	\begin{subfigure}[t]{0.3\linewidth}
		\centering
		\includegraphics[width=0.95\linewidth,trim={0 0 0 0},clip]{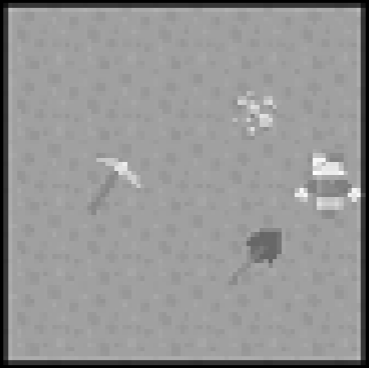}
		\caption{Ore type indistinguishable by agent's visual input.}
		\label{fig:sdworld_hard_gray}
	\end{subfigure}
	\caption{Mining domain. Ore type is indistinguishable by the grayscale visual observed by the agent.}
\end{figure} 

\paragraph{Attention Necessary to Learn in Some Domains} Temporal behaviors of the attention mechanism are also evaluated in a 2D Minecraft-like domain, where the agent must pick an appropriate tool (pickaxe or shovel) to mine either gold or iron ore (\Cref{fig:sdworld_hard_gold,fig:sdworld_hard_iron,fig:sdworld_hard_gray}). Critically, the agent observes identical images for both ore types, but unique audio features when near the ore, making long-term audio storage necessary for selection of the correct tool. The agent receives $+10$ reward for correct tool selection, $-10$ for incorrect selection, and $-1$ step cost. Compared to the door puzzle game, the mining domain is posed in such a way that the interplay between audio-video features is emphasized. Specifically, an optimal policy for this task must utilize both audio and video features: visual inputs enable detection of locations of the ore, agent, tools, whereas audio is used to identify the ore type.

\begin{figure}
	\centering		\includegraphics[width=1\linewidth]{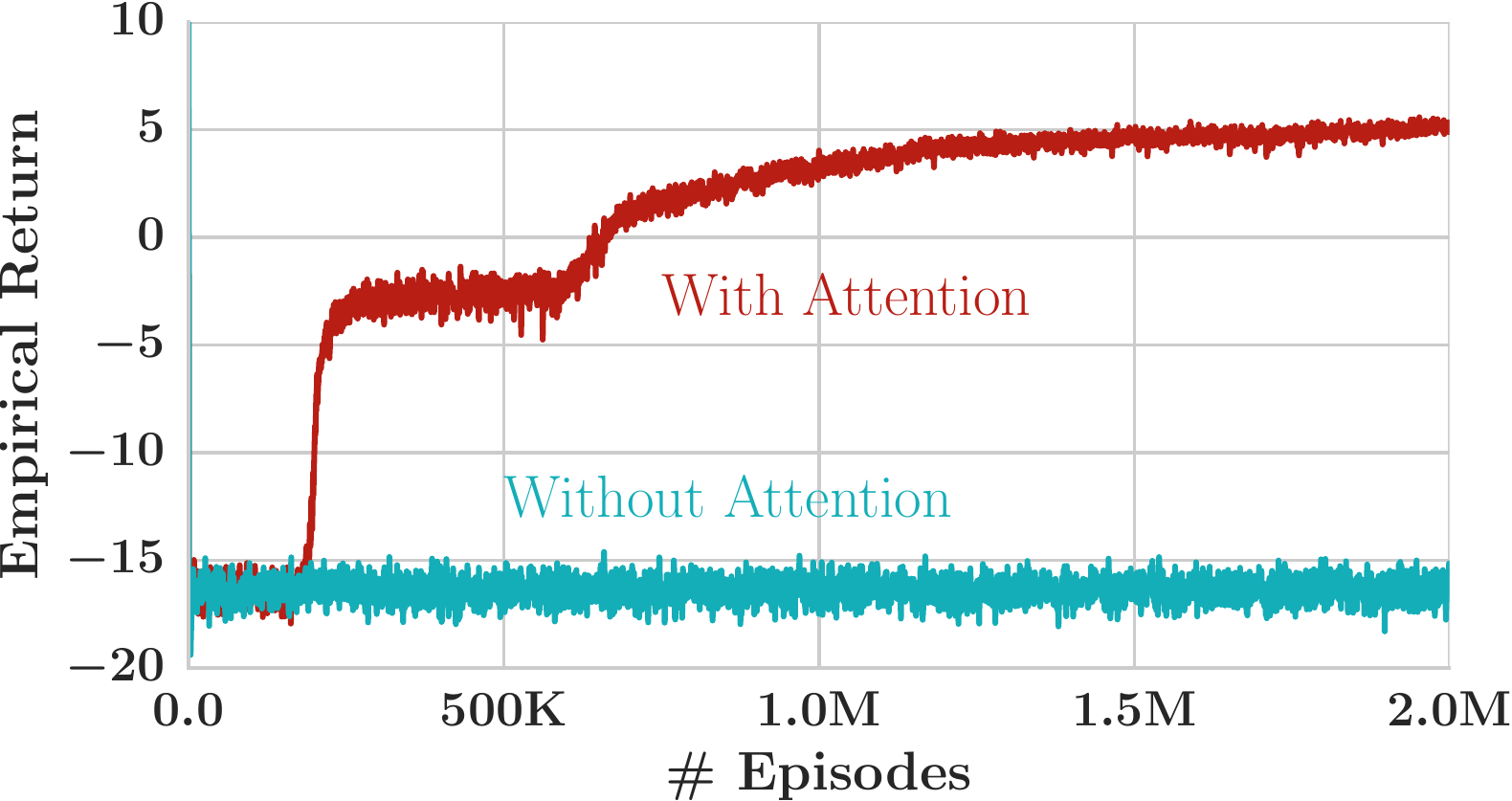}
	\caption{In the mining domain, the non-attentive network fails to learn, whereas the attentive network succeeds.}
	\label{fig:plot_sdworld_comparisons}
\end{figure}

\begin{figure*}[t]
	\centering
	\begin{subfigure}[t]{1\linewidth}
		\includegraphics[width=1\linewidth,trim={2.3cm 1.7cm 1.1cm 1.cm}, clip]{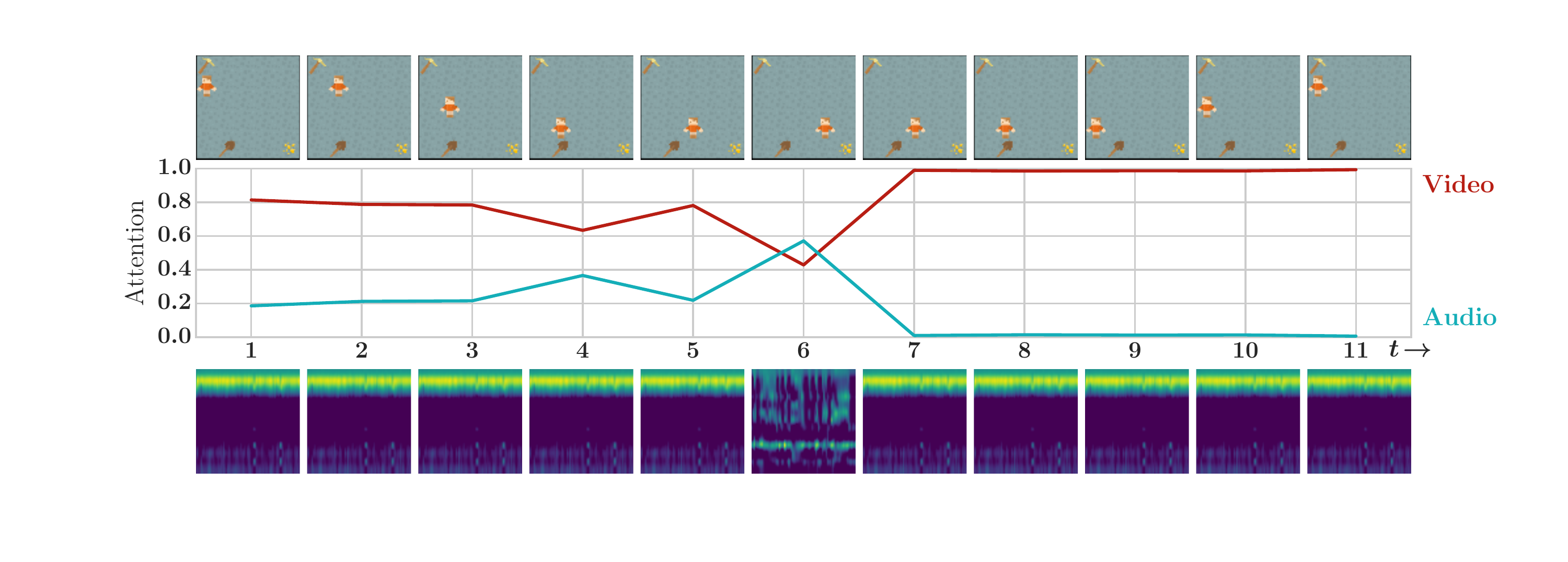}
		\caption{Agent anticipates salient audio features as it nears the ore, increasing audio attention until $t=6$. Audio attention goes to $0$ upon storage of ore indicator audio in the LSTM memory. Top and bottom rows show images and audio spectrogram sequences, respectively. Attention weights $\alpha^{t}$ plotted in center.}
		\label{fig:sdworld_attention}
	\end{subfigure}\\
	\begin{subfigure}[t]{1\linewidth}
		\includegraphics[width=1\linewidth,trim={2.3cm 0cm 1.1cm 0cm}, clip]{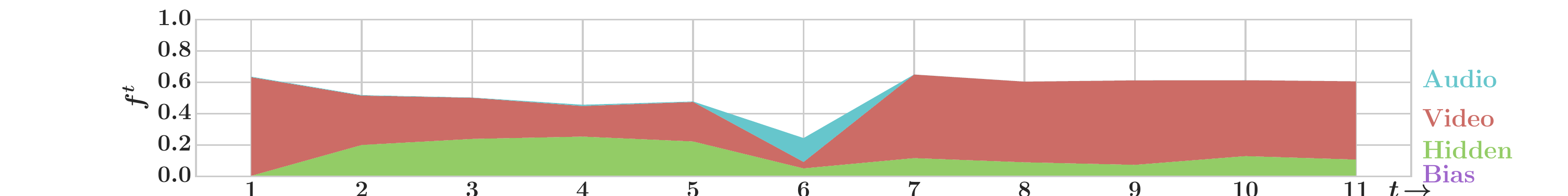}
		\caption{Average forget gate activation throughout episode. Recall $f^t = 0$ corresponds to complete forgetting of the previous cell state element.}
		\label{fig:sdworld_attention_lstm_forget}
	\end{subfigure}\\
	\begin{subfigure}[t]{1\linewidth}
		\includegraphics[width=1\linewidth,trim={2.3cm 0cm 1.1cm 0cm}, clip]{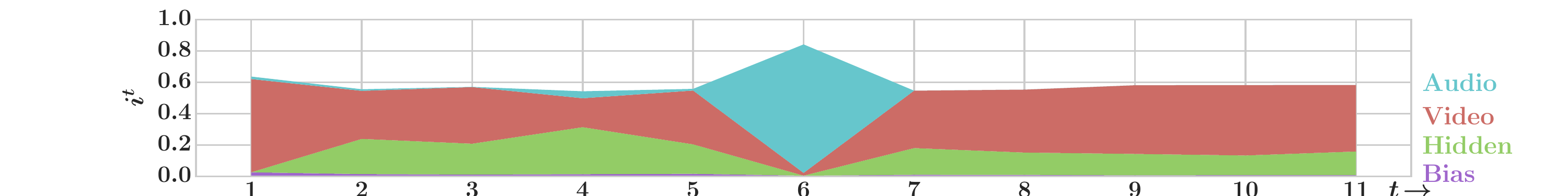}
		\caption{Average input gate activation throughout episode. Recall $i^t = 1$ corresponds to complete throughput of the corresponding input element.}
		\label{fig:sdworld_attention_lstm_input}
	\end{subfigure}
	\caption{Interactions of crossmodal attention and LSTM memory. At $t=6$, the attended audio input causes forget gate activation to drop, and the input gate activation to increase, indicating major overwriting of memory states. Relative contribution of audio to the LSTM forget and input activations drops to zero after the agent hears the necessary audio signal.}
\end{figure*}

Visual occlusion of the ore type, interplay of audio-video features, and sparse positive rewards cause the non-attentive network to fail to learn in the mining domain, as opposed to the attentive case (\Cref{fig:plot_sdworld_comparisons}). \Cref{fig:sdworld_attention} plots a sequence of frames where the agent anticipates salient audio features as it nears the ore at $t=6$, gradually increasing audio attention, then sharply reducing it to 0 after hearing the signal.

\subsection{Interactions of Attention and Memory}\label{sec:attention_lstm} 
While the anticipatory nature of crossmodal attention in the mining domain is interesting, it also points to additional lines of investigation regarding interactions of attention and updates of the agent's internal belief (as encoded in its LSTM cell state). Specifically, one might wonder whether it is necessary for the agent to place any attention on the non-useful audio signals prior to timestep $t=6$ in \Cref{fig:sdworld_attention}, and also whether this behavior implies inefficient usage of its finite-size memory state. 

Motivated by the above concerns, we conduct more detailed analysis of the interplay between the agent's attention and memory mechanisms as used in the CASL architecture (\Cref{fig:network_architecture}). We first provide a brief overview of LSTM networks to enable more rigorous discussion of these attention-memory interactions. At timestep $t$, LSTM cell state $C^t$ encodes the agent's memory given its previous stream of inputs. The cell state is updated as follows,
\begin{align}
f^t &= \sigma\left(W_f [\xatt,h^{t-1}] + b_f\right)\label{eqn:lstm_forget_gate_update},\\
i^t &= \sigma\left(W_i [\xatt,h^{t-1}] + b_i\right)\label{eqn:lstm_input_gate_update},\\
C^t &= f^t \odot C^{t-1} + i^t \odot \tanh\left(W_C[\xatt,h^{t-1}] + b_C\right)\label{eqn:lstm_cell_state_update},
\end{align}
where $f^t$ is the forget gate activation vector, $i^t$ is the input gate activation vector, $h^{t-1}$ is the previous hidden state vector, $\xatt$ is attended feature vector, and $\odot$ denotes the Hadamard product. Weights $W_f$, $W_i$, $W_C$ and biases $b_f$, $b_i$, $b_C$ are trainable parameters. The cell state update in \Cref{eqn:lstm_cell_state_update} first forgets certain elements ($f^t$ term), and then adds contributions from new inputs ($i^t$ term). Note that a forget gate activation of $0$ corresponds to complete forgetting of the previous cell state element, and that an input gate activation of $1$ corresponds to complete throughput of the corresponding input element. 

Our goal is to not only analyze the overall forget/input activations throughout the gameplay episode, but also to quantify the relative impact of each contributing variable (audio input, video input, hidden state, and bias term) to the overall activations. Many methods may be used for analysis of the contribution of explanatory variables in nonlinear models (i.e., \Cref{eqn:lstm_forget_gate_update,eqn:lstm_input_gate_update,eqn:lstm_cell_state_update}). We introduce a means of quantifying the correlation of each variable with respect to the corresponding activation function. In the following, we focus on the forget gate activation, but the same analysis applies to the input gate. First, expanding the definition of forget gate activation (\Cref{eqn:lstm_forget_gate_update}) assuming use of concatenated attention (\Cref{eqn:attn_attended_feat_concat}) yields,
\begin{align}
f^t = \sigma\left([W_{fa},W_{fv},W_{fh},b_f][\alpha_a x_a, \alpha_v x_v, h^{t-1}, I]\right),
\end{align}
where $x_a$ and $x_v$ are, respectively, the audio and video input features, and $I$ is the identity matrix. Define $\hat{f}^{t}_{m}$ as the forget gate activation if the $m$-th contributing variable were removed. For example, if audio input $x_a$ were to be removed, then,
\begin{align}
\hat{f}^{t}_{a} &= \sigma\left([W_{fv},W_{fh},b_f][\alpha_v x_v, h^{t-1}, I]\right).
\end{align}
Define the forget gate activation residual as $\tilde{f}^{t}_{m} = |f^t - \hat{f}^{t}_{m}|$ (i.e., the difference in output resulting from removal of the $m$-th contributing variable). Then, let us define a `pseudo correlation' of the $m$-th contributing variable with respect to the true activation,
\begin{align}
\rho(\tilde{f}^{t}_{m}) = \frac{\tilde{f}^{t}_{m}}{\sum_m \tilde{f}^{t}_{m}}\label{eqn:pseudo_correlation}.
\end{align}
This provides an approximate quantification of the relative contribution of the $m$-th variable (audio input, video input, hidden unit, or bias) to the overall activation of the forget and input gates.

Armed with this toolset, we now analyze the interplay between attention and LSTM memory. First, given the sequence of audio-video inputs in \Cref{fig:sdworld_attention}, we plot overall activations of the forget and input LSTM gates (averaged across all cell state elements), in \Cref{fig:sdworld_attention_lstm_forget} and \Cref{fig:sdworld_attention_lstm_input}, respectively. Critically, these plots also indicate the relative influence of each gate's contributing variables to the overall activation, as measured by \Cref{eqn:pseudo_correlation}. 

Interestingly, prior to timestep $t=6$, the contribution of audio to the forget gate and input gates is essentially zero, despite the positive attention on audio (in \Cref{fig:sdworld_attention}). At $t=6$, the forget gate activation drops suddenly, while the input gate experiences a sudden increase, indicating major overwriting of previous memory states with new information. The plots indicate that the attended audio input is the key contributing factor of both behaviors. In \Cref{fig:sdworld_attention}, after the agent hears the necessary audio signal, it moves attention entirely to video; the contribution of audio to the forget and input activations also drops to zero. Overall, this analysis indicates that the agent attends to audio in anticipation of an upcoming pertinent signal, but \emph{chooses} not to embed it into memory until the appropriate moment. Attention filters irrelevant sensor modalities, given the contextual clues provided by exogeneous and endogeneous input features; it, therefore, enables the LSTM gates to focus on learning when and how to update the agent's internal state.

\subsection{The Arcade Learning Environment}\label{sec:ale}
Preliminary evaluation of crossmodal attention was also conducted in the Arcade Learning Environment (ALE) \citep{bellemare13arcade}. We modified ALE to support audio queries, as it previously did not have this feature.
\begin{figure*}[t]
	\centering
	\includegraphics[width=1\linewidth,trim={0.cm 0.7cm 0.04cm 0.0cm}, clip]{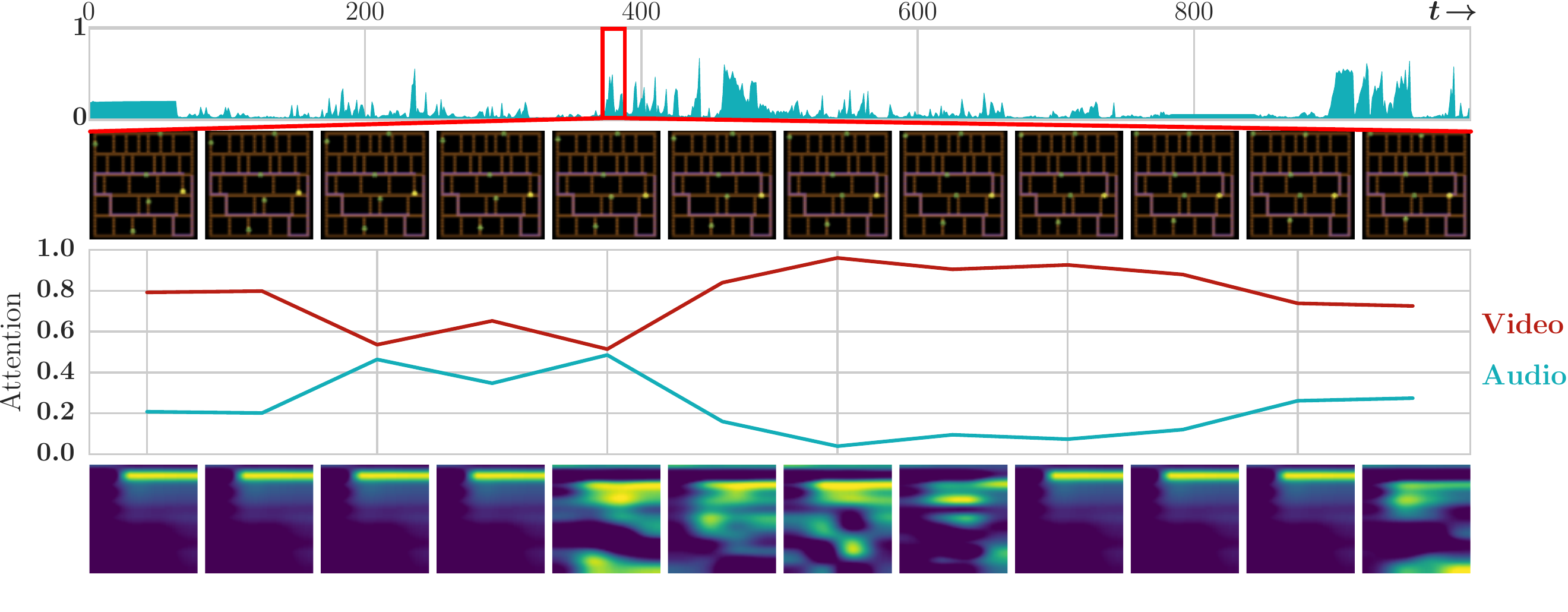}
	\caption{In Amidar, pathway vertices critical for avoiding enemies make an audible sound if not previously crossed. The agent anticipates and increases audio attention when near these vertices. Top row shows audio attention over 1000 frames, with audio/video/attention frames highlighted for zoomed-in region of interest. }
	\label{fig:amidar_attention}
\end{figure*}
\begin{table}[t]
	\caption{Preliminary results for learning in Atari 2600 games. The crossmodal attention learner, even \emph{without} options, achieves high score for non-hierarchical methods. We emphasize these are not direct comparisons due to our method leveraging additional sensory inputs, but are meant to highlight the performance benefits of crossmodal learning.}
	\label{table:amidar_results}
	\centering
	\setlength\tabcolsep{1.6pt} 
	\begin{tabular}{lcccc}
		\toprule
		Algorithm &  Hierarchical? & \multicolumn{1}{p{2cm}}{\centering Sensory\\ Inputs} & \multicolumn{1}{p{1cm}}{\centering Amidar\\ Score} & \multicolumn{1}{p{1cm}}{\centering H.E.R.O.\\ Score}\\
		\midrule
		DQN \cite{mnih2015human} &  \xmark & Video & 740 & 19950\\
		A3C \cite{mnih2016asynchronous} &  \xmark & Video & 284  & 28766 \\
		GA3C \cite{babaeizadeh2017ga3c} &  \xmark & Video & 218  & --\\
		Ours (\emph{no} options) &  \xmark & Audio \& Video & \fbox{900}  &  \fbox{32985}\\
		\midrule 
		A2OC \cite{harb2017waiting} &  \checkmark & Video & 880  & 20100 \\
		FeUdal \cite{vezhnevets2017feudal} &  \checkmark & Video & \textbf{$>$2500} & \textbf{$\sim$36000} \\
		\bottomrule
	\end{tabular}
\end{table}
Experiments were conducted in the Atari 2600 games H.E.R.O. and Amidar (\Cref{table:amidar_results}). This line of investigation considers impacts of crossmodal attention on Atari agent behavior, even without use of multiple (hierarchical) options; these results use CASL with a single option, hence tagged ``non-hierarchical" in the table.  Amidar was one of the games in which Deep Q-Networks failed to exceed human-level performance \citep{mnih2015human}. The objective in Amidar is to collect rewards in a rectilinear maze while avoiding patrolling enemies. The agent is rewarded for painting segments of the maze, killing enemies at opportune moments, or collecting bonuses. Background audio plays throughout the game, and specific audio signals play when the agent crosses previously-unseen segment vertices.  \Cref{fig:amidar_attention} reveals that the agent anticipates and increases audio attention when near these critical vertices, which are especially difficult to observe when the agent sprite is overlapping them (e.g., zoom into video sequences of \Cref{fig:amidar_attention}).

Our crossmodal attentive agent achieves a mean score of 900 in Amidar, over 30 test runs, outperforming the other non-hierarchical methods. A similar result is achieved for the game H.E.R.O., where our agent beats other non-hierarchical agents. Note agent also beats the score of the hierarchical approach of \cite{harb2017waiting}. We emphasize these are not direct comparisons due to our method leveraging additional sensory inputs, but are meant to highlight the performance benefits of crossmodal learning. We also note that the state-of-the-art hierarchical approach FeUdal \citep{vezhnevets2017feudal} beats our agent's score, and future investigation of the combination of audio-video attention with their approach may be of interest.
\section{Contribution}
This work introduced the Crossmodal Attentive Skill Learner (CASL), integrated with the recently-introduced Asynchronous Advantage Option-Critic (A2OC) architecture \citep{harb2017waiting} to enable hierarchical reinforcement learning across \emph{multiple} sensory inputs. 
We provided concrete examples where CASL not only improves performance in a single task, but accelerates transfer to new tasks. 
We demonstrated the learned attention mechanism anticipates and identifies useful sensory features, while filtering irrelevant sensor modalities during execution. 
We modified the Arcade Learning Environment \citep{bellemare13arcade} to support audio queries, and evaluations of crossmodal learning were conducted in the Atari 2600 games H.E.R.O. and Amidar.
Finally, building on the recent work of \citet{babaeizadeh2017ga3c}, we open-source a fast hybrid CPU-GPU implementation of CASL.
This investigation indicates crossmodal skill learning as a promising avenue for future works in HRL that target domains with high-dimensional, multimodal inputs.     
\section*{Acknowledgements}
This work was supported by Boeing Research \& Technology, ONR MURI Grant N000141110688, and BRC Grant N000141712072. 


\bibliographystyle{ACM-Reference-Format}  
\bibliography{./references}

\end{document}